# A Simple Model of Unbounded Evolutionary Versatility as a Largest-Scale Trend in Organismal Evolution

*Peter D. Turney*

*Institute for Information Technology*

*National Research Council of Canada*

*Ottawa, Ontario, Canada, K1A 0R6*

*peter.turney@iit.nrc.ca*

*Phone: 613-993-8564*

*Fax: 613-952-7151*

## Abstract

The idea that there are any large-scale trends in the evolution of biological organisms is highly controversial. It is commonly believed, for example, that there is a large-scale trend in evolution towards increasing complexity, but empirical and theoretical arguments undermine this belief. Natural selection results in organisms that are well adapted to their local environments, but it is not clear how *local* adaptation can produce a *global* trend. In this paper, I present a simple computational model, in which local adaptation to a randomly changing environment results in a global trend towards increasing evolutionary versatility. In this model, for evolutionary versatility to increase without bound, the environment must be highly dynamic. The model also shows that unbounded evolutionary versatility implies an accelerating evolutionary pace. I believe that unbounded increase in evolutionary versatility is a large-scale trend in evolution. I discuss some of the testable predictions about organismal evolution that are suggested by the model.

**Keywords:** evolutionary trends, evolutionary progress, large-scale trends, evolutionary versatility, evolvability, Baldwin effect.

**Running head:** Unbounded Evolutionary Versatility.

# A Simple Model of Unbounded Evolutionary Versatility as a Largest-Scale Trend in Organismal Evolution

## 1. Introduction

Ruse argues that almost all evolutionary theorists (before, after, and including Darwin) believe that there is progress in evolution [26]. Progress implies that there is a large-scale trend and that the trend is good [4, 5]. For example, it is commonly believed by the layperson that there is a large-scale trend in evolution towards increasing intelligence, and that this trend is good. Several scientists have suggested that we should focus on the (scientific) question of whether there are any large-scale trends, without regard to the (non-scientific) question of whether such trends are good [4, 5, 21, 22]. McShea presents an excellent survey of eight serious candidates ("live hypotheses") for large-scale trends in evolution: entropy, energy intensiveness, evolutionary versatility, developmental depth, structural depth, adaptedness, size, and complexity [21]. Complexity appears to be the most popular candidate.

The standard objection to large-scale trends in evolution is that natural selection is a *local* process that results in organisms that are well adapted to their *local* environments, and there is no way for this local mechanism to yield a *global* trend. On the other hand, it does seem that complexity (for example) has increased steadily since life on earth began. This seems to suggest that natural selection favours increasing complexity. However, many evolutionary theorists deny that there is any driving force, such as natural selection, behind any of the apparent large-scale trends in evolution.

Gould has presented the most extensive arguments against a driving force [16, 17]. Gould admits that there may be large-scale trends in evolution, but he argues that any such trends are, in essence, statistical artifacts. For example, if we consider the evolution of life since the first appearance of prokaryotes, the mean level of complexity would necessarily increase with time, because any organism with significantly less complexity than a prokaryote would not be able to live [16, 17]. According to Gould, the apparent trend towards increasing com-plexity is due to random variation in complexity plus the existence of a minimum level of complexity required to sustain life; there is no selective pressure that drives life towards increasing complexity. I discuss Gould's arguments in more detail in Section 2.

Of the eight live hypotheses for large-scale trends in evolution, this paper focuses on evolutionary versatility. I believe that there is indeed a selective advantage to increasing evolutionary versatility. Evolutionary versatility is the number of independent dimensions along which variation can occur in evolution [21, 25, 34, 35, 36, 37]. It is possible that increasing evolutionary versatility may be the driving force behind other apparent evolutionary trends, such as increasing complexity. I discuss the concept of evolutionary versatility in Section 3.

In Section 4, I introduce a simple computational model of unbounded evolutionary versatility. As far as I know, this is the first computational model of an evolutionary mechanism for one of the eight live hypotheses for large-scale trends in evolution. In this model, the population evolves in a series of *eras*. During each era, the fitness landscape is constant, but it randomly changes from one era to the next era. The model shows that there is a long-term trend towards increasing evolutionary versatility, *in spite of* the random drift of the fitness landscape. In fact, when the fitness landscape is constant, evolutionary versatility is bounded. In this model, unbounded evolutionary versatility *requires* a dynamic fitness landscape. The point of this model is to show that it is possible, in principle, for natural selection to drive evolution towards *globally* increasing evolutionary versatility, without bound, even though natural selection is a purely *local* process.

I discuss some related work in Section 5. My simple computational model of evolutionary versatility is related to Bedau and Seymour's model of the adaptation of mutation rates [8]. The primary focus of Bedau and Seymour was the adaptation of mutation rates, but the primary focus of this paper is evolutionary versatility. Bedau and Seymour's model does not address evolutionary versatility.

The core of this paper is the experimental evaluation of the model, in Section 6. In the first two experiments, I show that there are parameter settings for which evolutionary versatility can increase indefinitely. In the third experiment, I show that evolutionary versatility is bounded when the fitness landscape is static. In the remaining experiments, I examine a wide range of settings for the parameters in the model. These experiments show that the behaviour of the model is primarily determined by the parameters that control the amount of change in the fitness landscape.

In Section 7, I discuss the implications of the model. One of the most interesting implications of the model is that increasing evolutionary versatility implies an accelerating evolutionary pace. This leads to testable predictions about organismal evolution.

I discuss limitations and future work in Section 8 and I conclude in Section 9.

## 2. Arguments Against Large-Scale Trends in Evolution

Natural selection produces organisms that are well adapted to their local environments. The major objection to large-scale trends in evolution is that there is no way for local adaptation to cause a large-scale trend [16, 17, 22]. For example, although the environments of primates may favour increasing complexity, the environments of most parasites favour streamlining and simplification [17]. There is no generally accepted theoretical explanation of how natural selection could cause a large-scale trend. Van Valen's Red Queen hypothesis [33] attempts to explain how natural selection could cause a trend towards increasing complexity (based on coevolution), although Van Valen's hypothesis has been criticized [19, 23]. In this paper, I propose an explanation of how natural selection could cause a trend towards increasing evolutionary versatility. Attractive features of my proposal are that it can easily be simulated on a computer and that it leads to testable (in principle) predictions.

If there were some constant property, shared by all environments, then it would be easy to see how there could be a large-scale trend, due to long-term adaptation to this constant property. However, the computational model in Section 4 shows that there can be a large-scale trend even when the fitness landscape changes completely randomly over time.

Aside from theoretical difficulties with large-scale trends, there is the question of whether there is empirical evidence for any large-scale trend. McShea's survey of candidates for large-scale trends does not address the issue of evidence for the candidates [21], but, in another paper, he finds that there is no solid evidence for a trend in most kinds of complexity [20].

Gould argues that, even if there were empirical evidence for a large-scale trend, that does not imply that there is a driving force behind the trend [16, 17]. Gould argues that evolution is performing a random walk in complexity space, but there is a constraint on the minimum level of complexity. When the complexity of an organism drops below a certain level (e.g., the level of prokaryotes), it can no longer live. Gould's metaphor is that evolution is a drunkard's random walk, but with a wall in the way (i.e., a bounded diffusion process). This wall of minimum complexity causes random drift towards higher complexity. This random drift does not involve any active selection for complexity; there is no *push* or *drive* towards increased complexity.

In summary, (1) it is not clear how local selection can produce a global trend and (2) observation of a global trend does not imply that there is a driving force behind the trend. However, (1) my model shows one way in which local selection can produce a global trend and (2) the model makes testable (in principle) predictions.

## 3. Evolutionary Versatility

Evolutionary versatility is the number of independent dimensions along which variation can occur in evolution [21, 25, 34, 35, 36, 37]. A species with high evolutionary versatility has a wide range of ways in which it can adapt to its environment. Vermeij has argued that there should be selection for increased evolutionary versatility, because it can lead to organisms that are more efficient and better at exploiting their environments [34, 35, 36, 37].

An important point is that evolutionary versatility requires not merely many dimensions along which variation can occur, but also that the dimensions should be independent. *Pleiot-* *ropy* is the condition in which a single gene affects two or more distinct traits that appear to be unrelated. When *N* traits, which appear to vary on *N* dimensions, are linked by pleiotropy, there is effectively only one dimension along which variation can occur. Several authors have suggested that it would be beneficial for the genotype-phenotype map to be modular, since increasing modularity implies increasing independence of traits [2, 27, 30, 38]. McShea points out the close connection between evolutionary versatility and modularity [21].

Evolutionary versatility seems to be connected to several of the seven other "live hypotheses" [21]. Increasing evolutionary versatility implies increasing complexity, since the organisms must have some new physical structures to support each new dimension of variation. The dimensions are supposed to be independent, so the new physical structures must also be (at least partially) independent. The increasing accumulation of many independent new physical structures implies increasingly complex organisms. Among the other live hypotheses, developmental depth, structural depth, adaptedness, and perhaps energy intensiveness may be connected to evolutionary versatility [21].

Evolutionary versatility also seems to be related to evolvability [1, 2, 12, 13, 32, 38]. Evolvability is the capacity to evolve [12, 13]. An increasing number of independent dimensions along which variation can occur in evolution implies an increasing capacity to evolve, so it would seem that any increase in evolutionary versatility must also be an increase in evolvability. On the other hand, some properties that increase evolvability may decrease evolutionary versatility. For example, selection can be expected to favour a constraint that produces symmetrical left-right development [12, 13]. For humans, if a sixth finger were a useful mutation, then it would likely be best if the new fingers appeared simultaneously on both hands, instead of requiring two separate mutations, one for the left hand and another for the right hand. In general, selection should favour any constraint that produces adaptive covariation [24]. Such constraints increase evolvability, but they appear to decrease evolutionary versatility [21].

Increasing evolutionary versatility suggests an increasing number of independent dimensions, but adaptive covariation suggests a decreasing number of independent dimensions. Vermeij reconciles these forces by proposing that increasing evolutionary versatility adds more dimensions, which are then integrated by adaptive covariation, so that new dimensions are added and integrated in an ongoing cycle [21, 37].

Evolutionary versatility also appears to be related to the Baldwin effect [3, 18, 31]. The Baldwin effect is based on phenotypic plasticity, the ability of an organism (the phenotype) to adapt to its local environment, during its lifetime. Examples of phenotypic plasticity include the ability of humans to tan on exposure to sunlight and the ability of many animals to learn from experience. Phenotypic plasticity can facilitate evolution by enabling an organism to benefit from (or at least survive) a partially successful mutation, which otherwise (in the absence of phenotypic plasticity) might be detrimental. This gives evolution the opportunity to complete the partially successful mutation in future generations.

Evolution is not really free to vary along a given dimension if all variation along that dimension leads to death without children. Thus phenotypic plasticity increases the effective number of dimensions along which variation can occur in evolution. The Baldwin effect can therefore be seen as a mechanism for increasing evolutionary versatility.

## 4. A Simple Computational Model of Evolutionary Versatility

The following simple model of unbounded evolutionary versatility has three important features: (1) The fitness function is based on a shifting target, to demonstrate that a large-scale trend is possible, even when the optimal phenotype varies with time. In fact, in this model, the target *must* shift, if the model is to display unbounded evolutionary versatility. (2) The length of the genome can change. There is no upper limit on the possible length of the genome. This is necessary, because if the length were bounded, then there would be a finite number of possible genotypes, and thus there would be a bound on the evolutionary versatility. (3) The mutation rate is encoded in the genome, so that the mutation rate can adapt to the environment. This allows the model to address the claim that mutation becomes increasingly harmful as the length of the genome increases. Some authors have argued that natural selection should tend to drive mutation rates to zero [42]. Of course, if the mutation rate goes to zero, this sets a bound on evolutionary versatility.

Table 1 shows the parameters of the model and their baseline values. In the experiments that follow, I manipulate these parameters to determine their effects on the behaviour of the model. The meaning of the parameters in Table 1 should become clear as I describe the model.

Table 1: The parameters of the model with their baseline values.

|   | Parameter Name | Description | Baseline Value |
|---|---|---|---|
| 1 | POP_SIZE | number of individuals in population | 2000 |
| 2 | RUN_LENGTH | number of children born in one run of the model | 1000 |
| 3 | ERA_LENGTH | number of children born in one era | 100 |
| 4 | TARGET_CHANGE_RATE | fraction of target that changes between eras | 0.2 |
| 5 | TOURNAMENT_SIZE | number of individuals sampled when selecting parents | 400 |
| 6 | MUTATION_CODE_LENGTH | number of bits in genome for encoding the mutation rate | 10 |

Figure 1 is a pseudo-code description of the model of evolutionary versatility. In this model, a genome is a string of bits. The model is a *steady-state* genetic algorithm (as opposed to a *generational* genetic algorithm), in which children are born one-at-a-time [28, 29, 39, 40]. (In a generational genetic algorithm, the whole population is updated simultaneously, resulting in a sequence of distinct generations.) Parents are selected using tournament selection [9, 10]. In tournament selection, the population is randomly sampled and the two fittest individuals in the sample are chosen to be parents (see lines 15 to 17 in Figure 1). The selective pressure can be controlled by varying the size of the sample (TOURNAMENT_SIZE). A new child is created by applying single-point crossover to the parents (lines 18 to 21). The new child then undergoes mutation, based on a mutation rate that is encoded in the child's genome (lines 22 to 28). Mutation can flip a bit (from 0 to 1 or from 1 to 0) in the genome or it can add or delete a bit, making the bit string longer or shorter.

The initial section of a genome (the first MUTATION_CODE_LENGTH bits) encodes the mutation rate for that genome. The remainder of the genome (which may be null) encodes the phenotype. The phenotype is a bit string, created from the genome by simply copying the bits from the genotype, beginning with the MUTATION_CODE_LENGTH plus one bit of the genotype and continuing to the end of the genotype. If the length of the genome is exactly MUTATION_CODE_LENGTH (as it is when the simulation first starts running), then the phenotype is the null string.

The fitness of the phenotype is determined by comparing it to a target. The target is a random string of bits. The fitness of the phenotype is the number of matching bits between the phenotype and the target (lines 31 to 33). If the phenotype is null, the fitness is zero. The length of the target grows, so that the target is always at least as long as the longest phenotype in the population (lines 29 to 30). When a new child is born, if it is fitter than the least fit individual in the population, then it replaces the least fit individual (lines 34 to 36).

The target is held constant for an interval of time called an *era*. At the end of an era, the target is randomly changed. Each time the target changes, it is necessary to re-evaluate the fitness of every individual (lines 37 to 40). Instead of dividing a run into a series of eras, the model could have been designed to have a small, continuous change of the target for each new child that is born. (This is a special case of the current model, where TARGET_CHANGE_RATE is small and ERA_LENGTH is one.) The main motivation for dividing the run into a series of eras is to increase the computational efficiency of the model, since it is computationally expensive to re-evaluate the fitness of every individual each time a new child is born. (Actually, it would only really be necessary to re-evaluate TOURNAMENT_SIZE individuals each time a new child is born.) It could also be argued that organismal evolution is characterized by periods of stasis followed by rapid change (e.g., punctuated equilibria),

```
1    set the parameter values:
2        POP_SIZE                   — number of individuals in population
3        RUN_LENGTH                 — number of children born in one run
4        ERA_LENGTH                 — number of children born in one era
5        TARGET_CHANGE_RATE         — fraction of target that changes between eras
6        TOURNAMENT_SIZE            — number of individuals sampled when selecting parents
7        MUTATION_CODE_LENGTH       — number of bits in genome for encoding mutation rate
8    let Pop be an array of POP_SIZE bit strings;        — the population
9    let each bit string Pop[i] be a string of MUTATION_CODE_LENGTH randomly generated bits, where 0
10       and 1 are generated with equal probability;     — Pop[i] is the i-th individual in Pop
11   let Target be an empty string;               — the goal string for determining fitness
12   let Fit be an array of POP_SIZE integers;     — the fitness of Pop
13   let each Fit[i] be 0;                         — Fit[i] is the initial fitness for Pop[i]
14   for ChildNum = 1 to RUN_LENGTH do:           — main loop
15       randomly sample TOURNAMENT_SIZE individuals (bit strings) from Pop (sampling with
16           replacement) and take the two fittest individuals to be parents;
17       randomly let Mom be one of the two parents and let Dad be the other;
18       randomly pick a crossover point Cross that falls inside the bit strings of
19           both Mom and Dad;                     — the parents may have different lengths
20       let Child be the left side of Mom's bit string, up to Cross, followed by the right side of
21           Dad's bit string, after Cross;        — thus length(Child) equals length(Dad)
22       let Mutate be set to the Child's mutation rate (a fraction between 0 and 1) by
23           interpreting the first MUTATION_CODE_LENGTH bits of Child as an encoded
24           fraction;                             — example: '00' = 0, '01' = 1/3, '10' = 2/3, '11' = 1
25       randomly flip bits in Child, where the probability of flipping any bit is Mutate;
26       randomly add (remove) a bit to (from) the end of Child, with a probability of Mutate,
27           where adding and removing have equal probability, but do not remove a bit
28           if length(Child) = MUTATION_CODE_LENGTH;    — minimum required length
29       if length(Child) − MUTATION_CODE_LENGTH > length(Target), then randomly add
30           a bit to Target;                      — 0 or 1 with equal probability
31       let ChildFit be the number of bits in Child that match the bits in Target, where the
32           first bit in Target is aligned with the MUTATION_CODE_LENGTH + 1 bit of
33           Child;                                — if Child is too short, ChildFit is 0
34       let Worst be the oldest individual among the least fit individuals in Pop;
35       let WorstFit be the fitness of Worst;
36       if ChildFit > WorstFit, then replace Worst with Child and replace WorstFit with ChildFit;
37       if ERA_LENGTH divides into ChildNum with no remainder, then do:
38           randomly flip bits in Target, where the probability of flipping any
39               bit is TARGET_CHANGE_RATE;
40           re-evaluate the fitness Fit[i] of every individual Pop[i];
41       end if;
42   end for;
```

Figure 1: A pseudo-code description of the model of evolutionary versatility. The model is a steady-state genetic algorithm with crossover and mutation. Mutation can flip a bit in the genome or increase or decrease the genome length by one bit. The mutation rate is encoded in the genome. Parents are chosen by tournament selection.

so this feature of the model makes it more realistic.

Recall that evolutionary versatility is the number of independent dimensions along which variation can occur in evolution. In this model, the evolutionary versatility of a genome is the length of the genome minus MUTATION_CODE_LENGTH. This is the length of the part of the genome that encodes the phenotype. The first MUTATION_CODE_LENGTH bits are not independent and they do not directly affect the phenotype, so I shall ignore them when counting the number of independent dimensions along which variation can occur. Each remaining bit in the genome is an independent dimension along which variation can occur. The dimensions are independent because the fitness of the organism is defined as the number of matches between the phenotype and the target; that is, the fitness is the sum of the fitnesses for each dimension. Fitness on one dimension (a match on one bit) has no impact on fitness on another dimension (a match on another bit).

Note that increasing evolutionary versatility (i.e., increasing genome length) does not necessarily imply increasing fitness, because (1) the additional bits do not necessarily match the target and (2) a mutation rate that enables evolutionary versatility (genome length) to increase also makes the genome vulnerable to disruptive (fitness reducing) mutations. However, the design of the model implies that increasing genome length will tend to be *correlated* with increasing fitness.

## 5.  Related Models

The most closely related work is the model of Bedau and Seymour [8]. In Bedau and Seymour's model, mutation rates are allowed to adapt to the demands of the environment. They find that mutation rates adapt to an optimal level, which depends on the evolutionary demands of the environment for novelty. My model is similar, in that mutation rates are also allowed to adapt. Other work with adaptive mutation rates includes [6, 11, 14, 41]. Bedau and Seymour's model and my model are distinct from this other work in that we share an interest in the relationship between the adaptive mutation rates and the evolutionary demands of the environment for novelty.

The main difference between this paper and previous work is the different objective. None of the previous papers were concerned with large-scale trends in evolution. As far as I know, this is the first model to show how it is possible for evolutionary versatility to increase without bound.

## 6.  Results of Experiments with the Model

This section presents eight experiments with the model of evolutionary versatility. The first experiment examines the behaviour of the model with the baseline parameter settings. The second experiment runs the model for ten million births, but is otherwise the same as the baseline case. This experiment gives a lower resolution view of the behaviour of the model, but over a much longer time scale. These two experiments support the claim that the model can display unbounded evolutionary versatility, given suitable parameter settings. The third experiment uses the baseline parameter settings, except that the target is held constant. With a constant target, the mutation rate eventually goes to zero and the population becomes static. The results show that, in this model, unbounded evolutionary versatility requires a dynamically varying target. The remaining experiments vary the parameters of the model, one at a time. These experiments show that the model is most sensitive to the parameters that determine the pace of change in the target. In comparison, the parameters that do not affect the target have relatively little influence on the large-scale behaviour of the model.

### 6.1    Experiment 1: Baseline Parameter Values

Figure 2 shows the results with the baseline parameter settings (see Table 1). Since the model is stochastic, each run is different (assuming the random number seed is different), but the general behaviour is the same for all runs (assuming the parameters are the same). In this experiment, I ran the model 100 times and averaged the results across the 100 runs.

For this experiment, the length of an era is 100 children. At the start of each new era, the fitness drops. However, the overall trend is towards increasing fitness (see the first plot in Figure 2). Although the probability that a mutation will increase the genome length is equal to the probability that a mutation will decrease the genome length, there is a steady trend towards increasing genome length (the second plot in Figure 2). The mutation rate decreases steadily (third plot). Although the length of an era is fixed, in each era, the increase in fitness since the start of the era is greater than the corresponding increase for the previous era (fourth plot). This shows that the pace of evolution is accelerating.

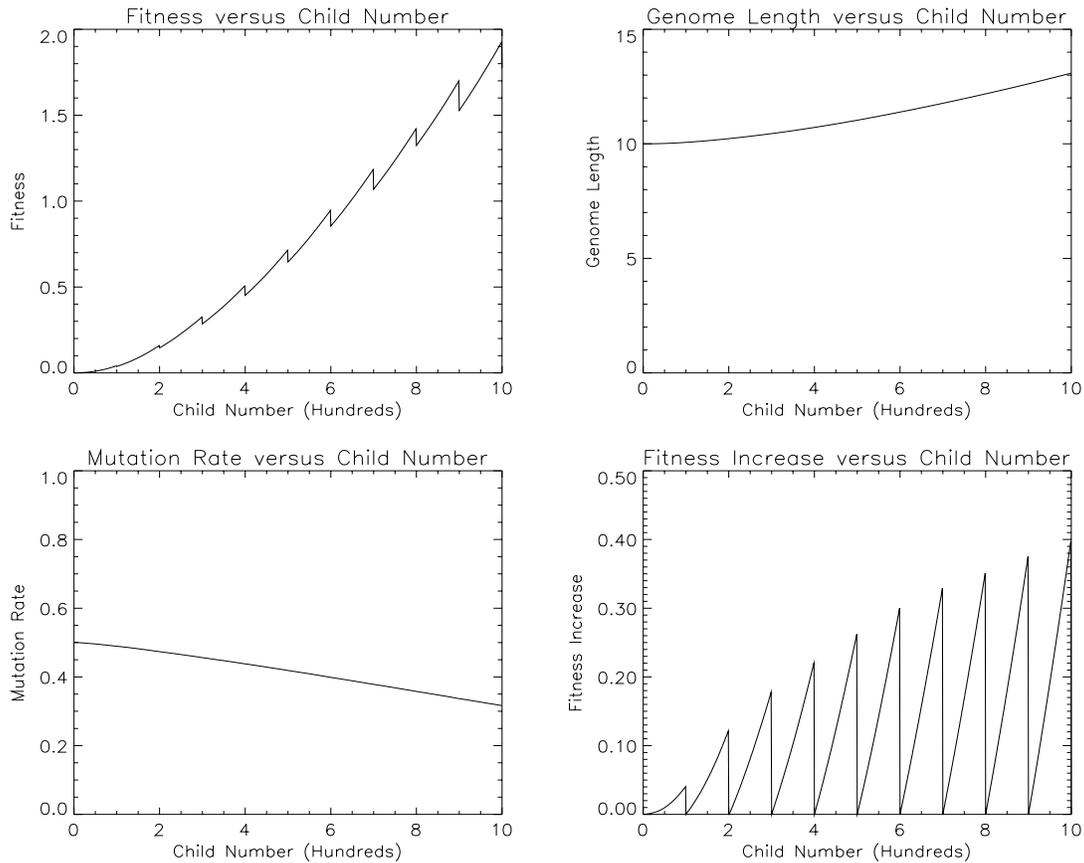

Figure 2: Experiment 1: Baseline parameter values. These four plots show the fitness, genome length, mutation rate, and fitness increase as functions of the number of children that have been born. The target for the fitness function changes each time one hundred children are born. The fitness increase is the increase in fitness since the most recent change in the target. All values are averages over the whole population, for one hundred separate runs of the baseline configuration (2,000 individuals times 100 runs yields 200,000 samples per value).

Evolutionary versatility is given by the genome length (minus MUTATION_CODE_LENGTH). The steady growth in the genome length (in the second plot in Figure 2) shows that evolutionary versatility is increasing, at least over the relatively short time span of this experiment.

## 6.2    Experiment 2: Longer Run Length

The steady decrease in the mutation rate in the first experiment suggests that the mutation rate might go to zero. If the mutation rate is zero, then the fitness can no longer increase without bound. The fitness would vary randomly up and down as the target changed each era, but the fitness would always be less than the genome length, which would become a constant value.

In the second experiment, I ran the model until 10,000,000 children were born (RUN_LENGTH = 10,000,000), in order to see whether the trends in Figure 2 would continue over a longer time scale. I ran the model 10 times and averaged the results across the 10 runs. In the first experiment, the population averages (for fitness, genome length, mutation rate, and fitness increase) were calculated each time a new child was born. In the second experiment, to increase the speed of the model, the population averages were only calculated each time 10,000 children were born. Figure 3 shows the results for the second experiment.

Figure 3 shows that the trends in Figure 2 continue, in spite of the much longer time scale. The only exception is the mutation rate, which quickly falls from its initial value of 0.5 to hover between 0.03 and 0.05. There is no indication that the mutation rate will go to zero. However, since the model is stochastic, there is always a very small (but non-zero) probability that the mutation rate could go to zero.

In Figure 3, the fitness increase is calculated as the average fitness of the population at the end of an era minus the average fitness of the population at the start of the same era. The fitness increase is calculated each era and then the average fitness increase is calculated for each 10,000 births. Since there are 100 births in an era, there are 100 eras in each sample of 10,000 births, so each value in the plot of the fitness increase is the average of 100 eras and 10 runs. The values in the other three plots (fitness, genome length, and mutation rate) are averages over 10 runs.

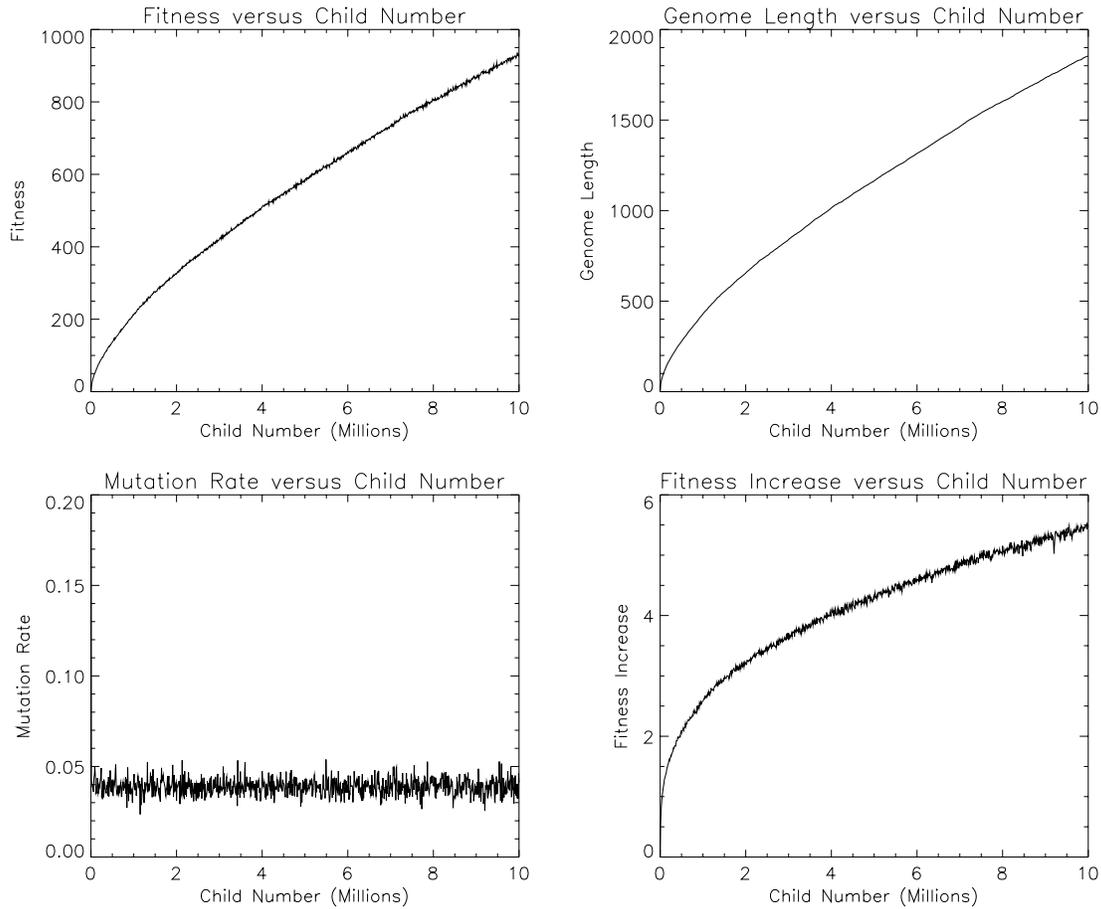

Figure 3: Experiment 2: Longer run length. These four plots show the fitness, genome length, mutation rate, and fitness increase as functions of the number of children that have been born. As in the first experiment, the target for the fitness function changes each time one hundred children are born. All values are averages over the whole population, for ten separate runs of the baseline configuration. The values are calculated once for each ten thousand children that are born.

### 6.3    Experiment 3: Static Target

This experiment investigated the behaviour of the model when the target was static. As in the second experiment, the population averages for fitness, genome length, and mutation rate were calculated once every 10,000 births. I ran the model 10 times and averaged the results across the 10 runs. I used the baseline parameter settings, except for RUN_LENGTH, ERA_LENGTH, and TARGET_CHANGE_RATE. I set both RUN_LENGTH and ERA_LENGTH to 10,000,000 and I set TARGET_CHANGE_RATE to zero. Figure 4 shows the results of the runs.

In all 10 runs, the mutation rate was zero, for every member of the population, long before 10,000,000 children were born. The longest run lasted for 349,000 births, the shortest run lasted for 37,100 births, and the average run lasted for 165,300 births. In comparison, in the second experiment, all 10 runs ran for 10,000,000 children, with no sign that the mutation rate would ever reach zero. These experiments support the claim that (in this model) unbounded evolutionary versatility requires a dynamic target. The following two experiments investigate the amount of change in the target that is needed to ensure unbounded evolutionary versatility.

### 6.4    Experiment 4: Varying Rate of Change of Target

In the fourth experiment, the rate of change of the target was varied from 0.0 to 0.2. The RUN_LENGTH was constant at 1,000,000. The remaining parameters were set to their baseline values. Figure 5 shows the behaviour of the model, averaged over ten separate runs. The time of the birth of the last novel child (i.e., the time at which the mutation rate becomes zero for every member of the population) was around the birth of the 100,000th child when TARGET_CHANGE_RATE was 0.0, but it quickly rose to around the 1,000,000th child as TARGET_CHANGE_RATE approached 0.1 (see the first plot in Figure 5). It could not go past 1,000,000, because RUN_LENGTH was 1,000,000. I conjecture that there is a threshold for TARGET_CHANGE_RATE at approximately 0.1, where the average time of birth of the last novel child approaches infinity as RUN_LENGTH approaches infinity.

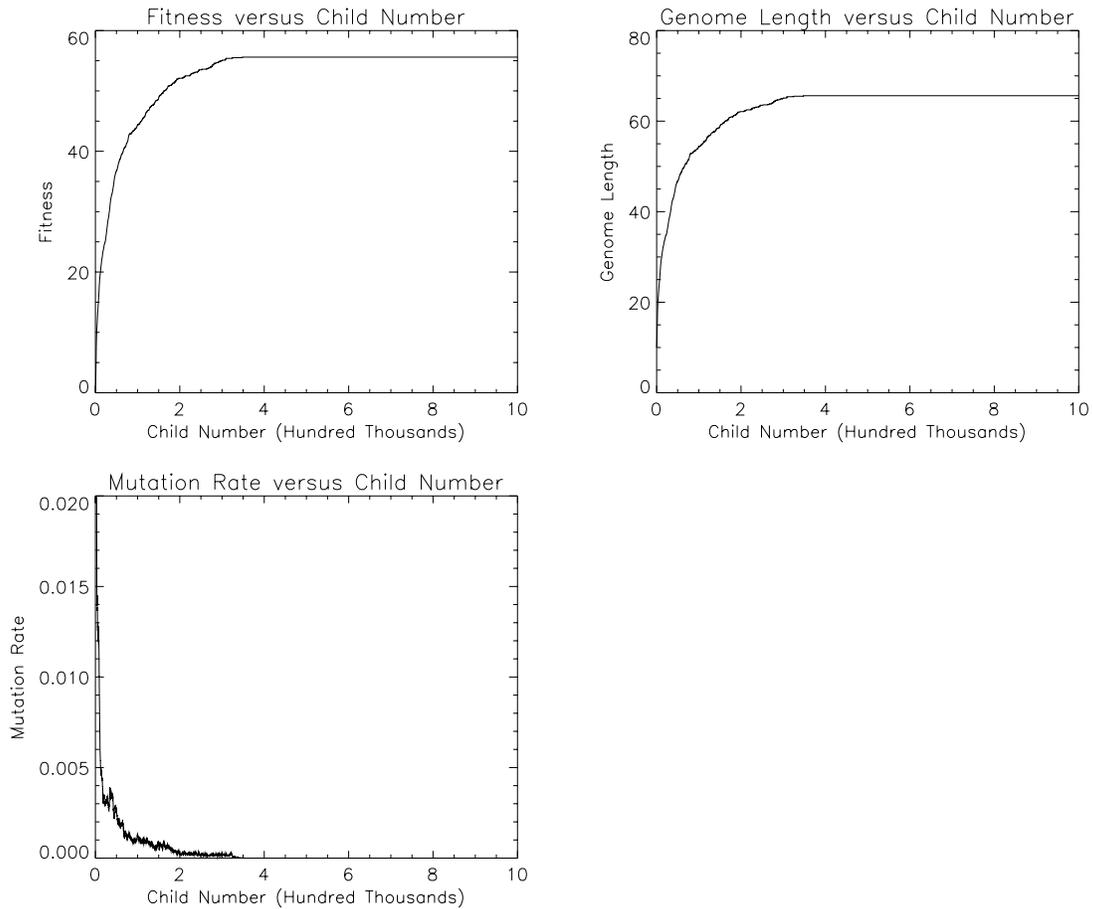

Figure 4: Experiment 3: Static target. These three plots show the fitness, genome length, and mutation rate as functions of the number of children that have been born. Since the target is static, the fitness increase is undefined. All values are averages over the whole population, for ten separate runs of the model. The values are calculated once for each ten thousand children that are born.

When TARGET_CHANGE_RATE was 0.08, 40% of the ten runs made it all the way to the 1,000,000th birth with a mutation rate above zero. When the TARGET_CHANGE_RATE was 0.1, this went to 90% (second plot in Figure 5). The average fitness of the population at the time of the birth of the 1,000,000th child (the final fitness) rose steadily as TARGET_CHANGE_RATE increased from 0.0 to 0.1 (third plot). Above 0.1, it could not rise significantly, because of the limit set by RUN_LENGTH. I conjecture that it would rise to infin-ity as RUN_LENGTH rises to infinity. The average mutation rate of the population at the time of the birth of the 1,000,000th child (the final mutation rate) increased steadily as TARGET_CHANGE_RATE increased, even past the 0.1 threshold.

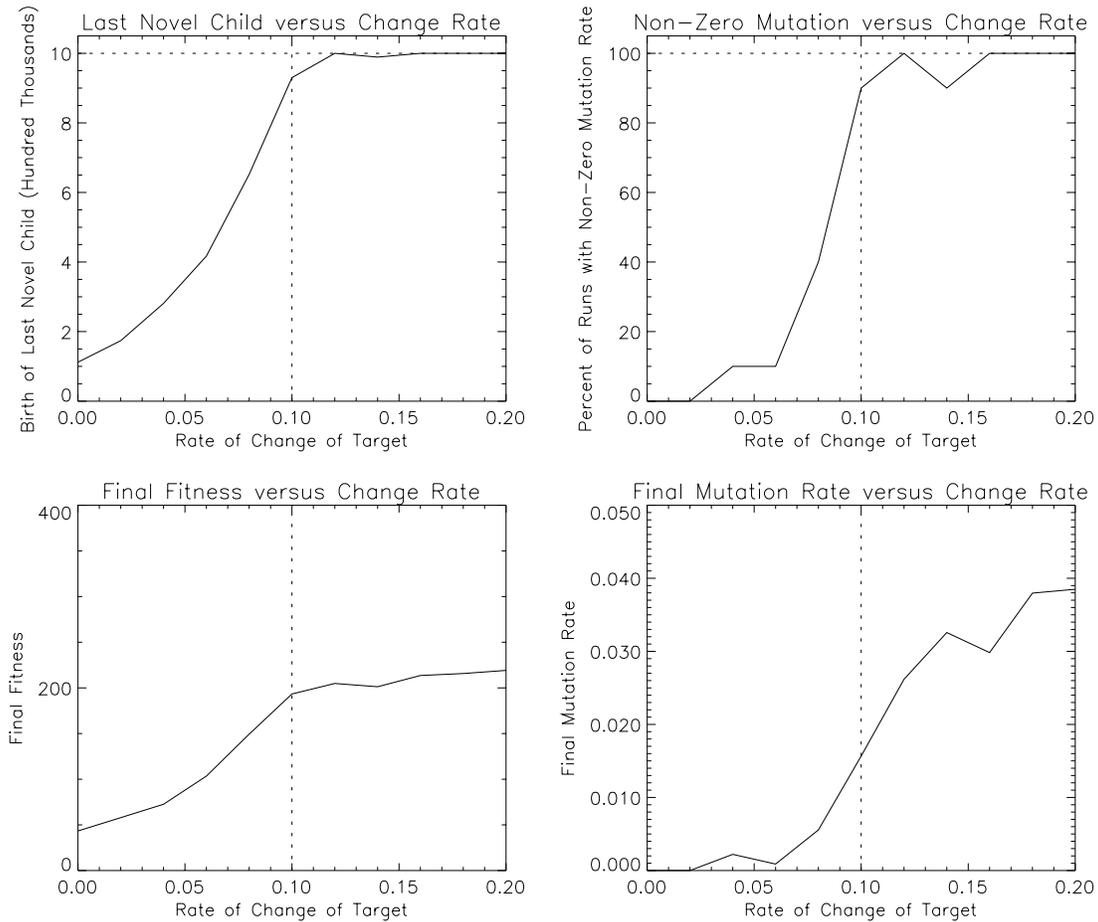

Figure 5: Experiment 4: Varying rate of change of target. In this experiment, TARGET_CHANGE_RATE varies from 0.0 (its value in Experiment 3) to 0.2 (its value in Experiments 1 and 2). The RUN_LENGTH is 1,000,000. When TARGET_CHANGE_RATE is about 0.1, there is a qualitative change in the behaviour of the model. This threshold appears to separate bounded evolutionary versatility (as in Experiment 3; left of the vertical dotted line) from unbounded evolutionary versatility (as in Experiment 2; right of the vertical dotted line). All values in the plots are based on ten separate runs of the model.

This experiment suggests that a relatively high amount of change is required to ensure that evolutionary versatility will increase without bound. When the target is changed once every hundred children (ERA_LENGTH = 100), the target must change by at least 10% (TARGET_CHANGE_RATE = 0.1). If there is less environmental change than this, the mutation rate eventually drops to zero.

### 6.5    Experiment 5: Varying Length of Era

In the fifth experiment, the length of an era was varied from 100 to 1000. The RUN_LENGTH was constant at 1,000,000. The remaining parameters were set to their baseline values. Figure 6 shows the behaviour of the model, averaged over ten separate runs.

Like Experiment 4, this experiment supports the hypothesis that a relatively high amount of change is required to ensure that evolutionary versatility will increase without bound. When the target changes by 20% each era (TARGET_CHANGE_RATE = 0.2), the length of an era cannot be more than 200 children (ERA_LENGTH = 200), if the mutation rate is to stay above zero.

### 6.6    Experiment 6: Varying Tournament Size

In the sixth experiment, the tournament size was varied from 100 to 1000. The baseline value for TOURNAMENT_SIZE was 400. Larger tournaments mean that there is more competition to become a parent, so there is higher selective pressure. The RUN_LENGTH was constant at 1,000,000 and the remaining parameters were set to their baseline values. Figure 7 shows the behaviour of the model, averaged over ten separate runs.

The results suggest that the model will display unbounded evolutionary versatility as long as TOURNAMENT_SIZE is more than about 200. Compared to ERA_LENGTH and TARGET_CHANGE_RATE, the behaviour of the model is relatively robust with respect to TOURNAMENT_SIZE. The model displays unbounded evolutionary versatility for a relatively wide range of values of TOURNAMENT_SIZE.

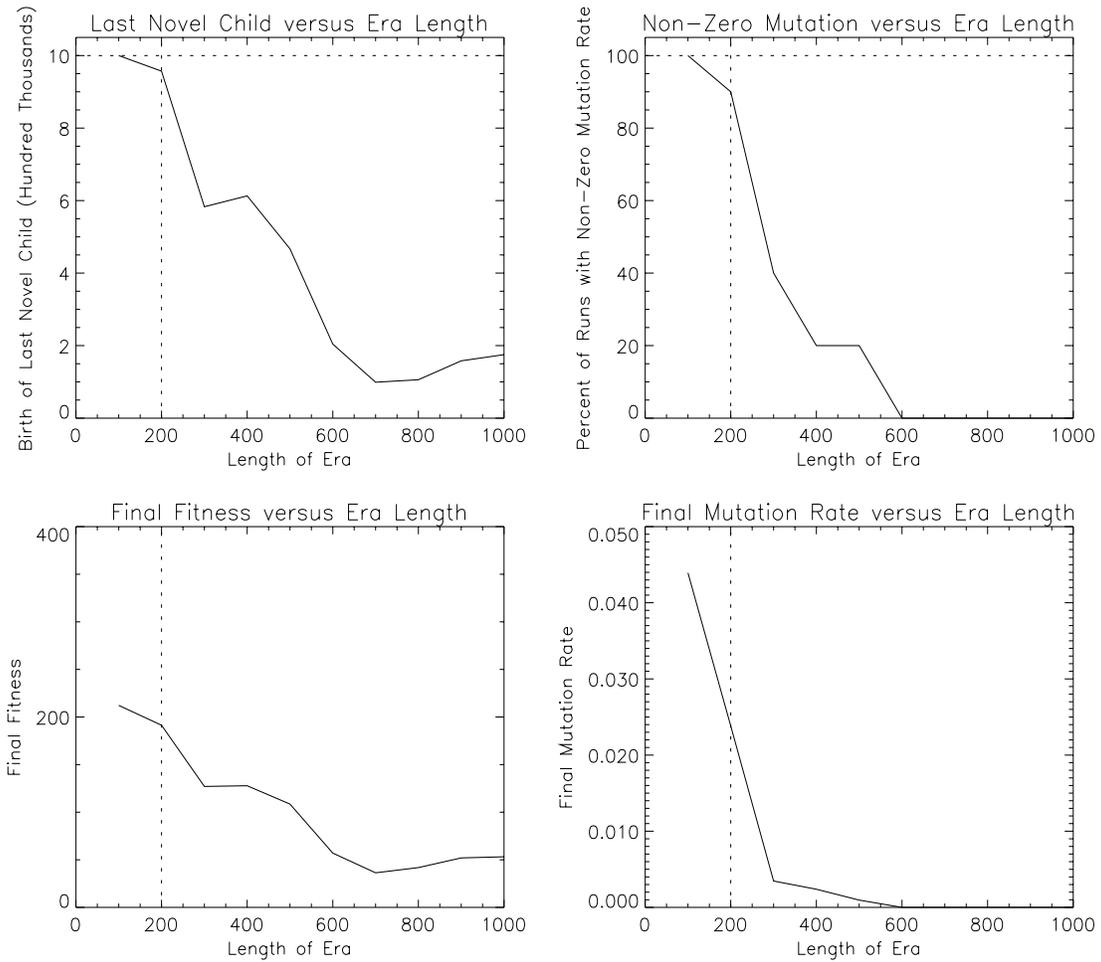

Figure 6: Experiment 5: Varying length of era. In this experiment, ERA_LENGTH varies from 100 (its value in Experiments 1 and 2) to 1000 (its length in Experiment 3 was 10,000,000). The RUN_LENGTH is 1,000,000. When ERA_LENGTH is about 200, there is a qualitative change in the behaviour of the model. This threshold appears to separate unbounded evolutionary versatility (as in Experiment 2; left of the vertical dotted line) from bounded evolutionary versatility (as in Experiment 3; right of the vertical dotted line). All values in the plots are based on ten separate runs of the model.

### 6.7 Experiment 7: Varying Population Size

In the seventh experiment, the size of the population was varied from 1000 to 3000. The baseline value of POPULATION_SIZE was 2000. The RUN_LENGTH was constant at 1,000,000 and the remaining parameters were set to their baseline values. Figure 8 shows the behaviour

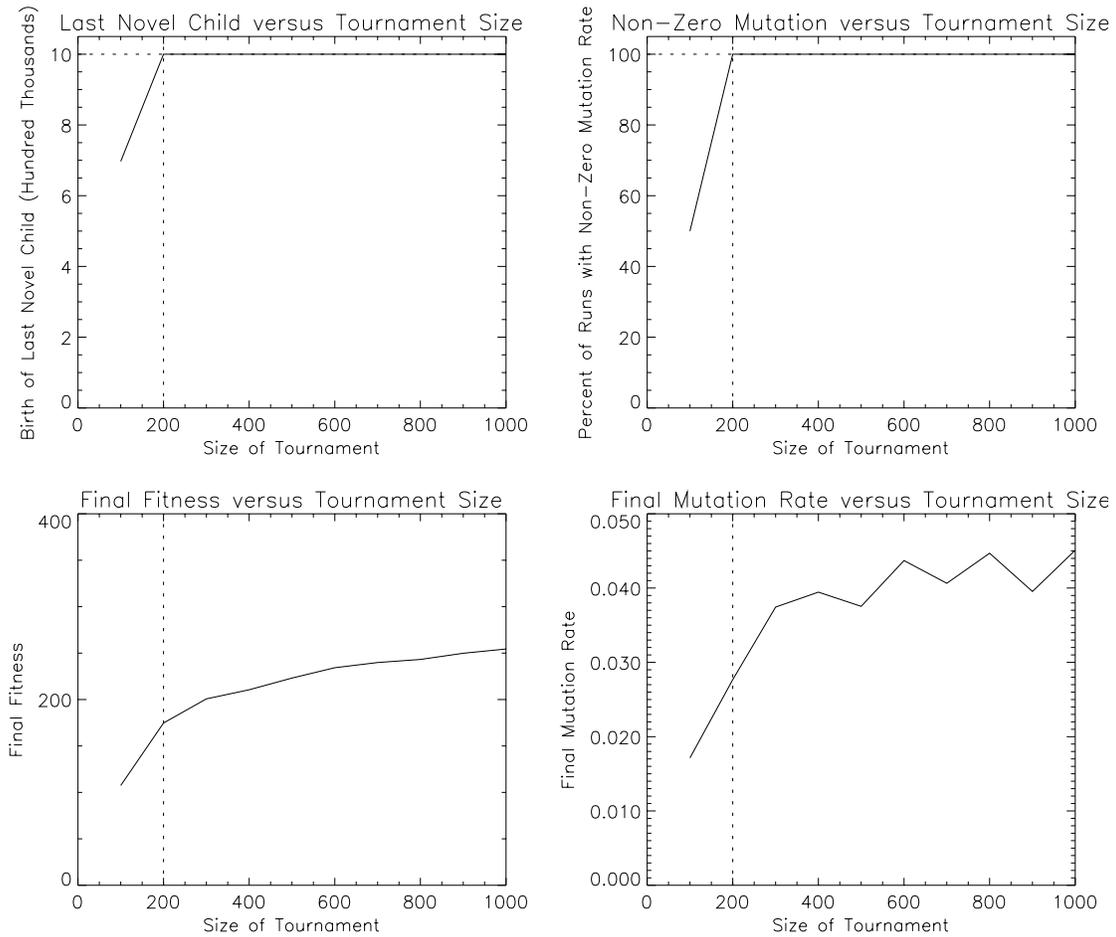

Figure 7: Experiment 6: Varying tournament size. In this experiment, TOURNAMENT_SIZE varies from 100 to 1000. The RUN_LENGTH is 1,000,000. Larger tournaments mean greater selective pressure. The results suggest that there is unbounded evolutionary versatility as long as TOURNAMENT_SIZE is greater than about 200 (see the first two plots). In the third plot, the final fitness (the average fitness of the population at the time of the birth of the 1,000,000th child) continues to rise even when TOURNAMENT_SIZE is greater than 200 and 100% of the runs reach the 1,000,000th child with a non-zero mutation rate (see the second plot). This suggests that there is an advantage to higher selective pressure, beyond what is needed to obtain unbounded evolutionary versatility.

of the model, averaged over ten separate runs.

When the population is small, the model is more susceptible to random variations. With a large population, the model will tend to behave the same way, every time it runs. Figure 8 suggests that the model becomes unstable when the population size is less than about 2000, although there is no sharp boundary at 2000. This is unlike Experiment 4, where there is a sharp boundary when TARGET_CHANGE_RATE is 0.1, and Experiment 5, where there is a sharp boundary when ERA_LENGTH is 200.

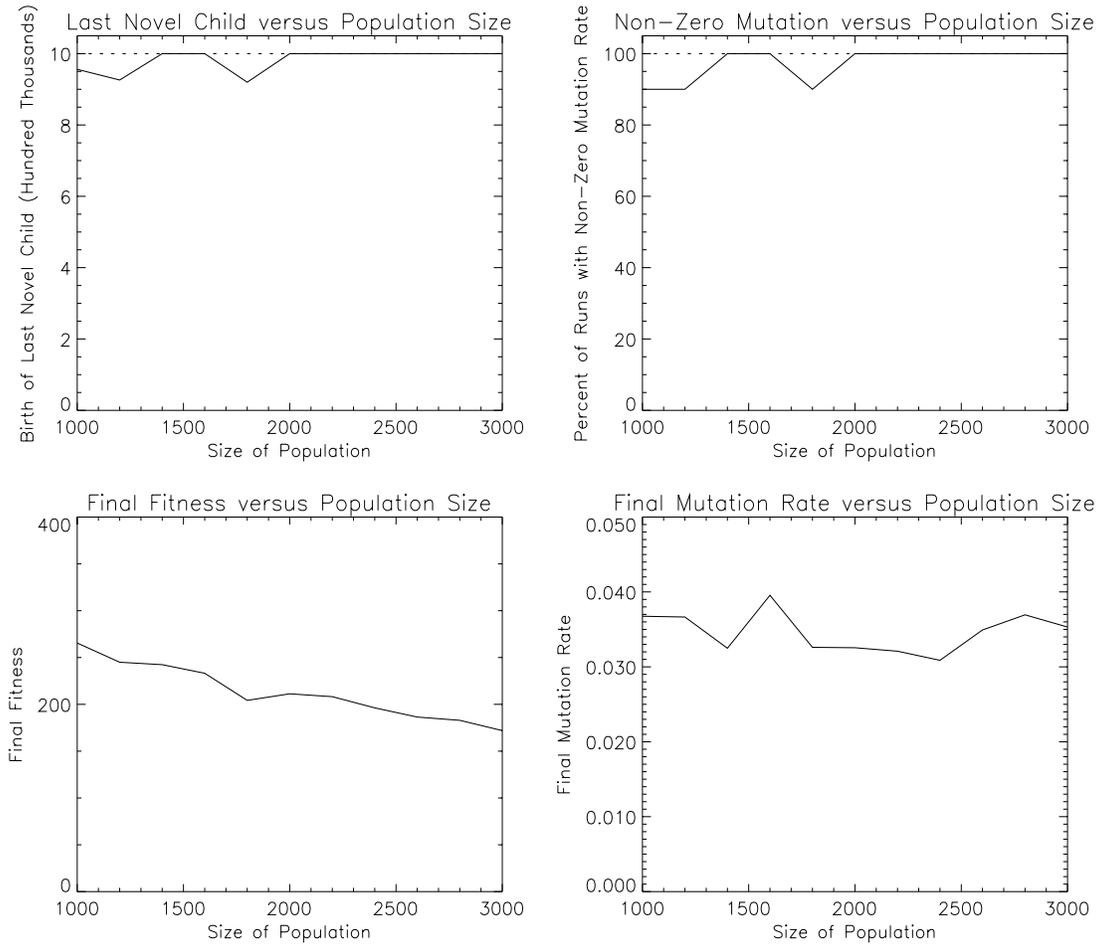

Figure 8: Experiment 7: Varying population size. In this experiment, POPULATION_SIZE is varied from 1000 to 3000. The baseline value of POPULATION_SIZE is 2000. These plots suggest that, in most runs, we will have unbounded evolutionary versatility, even when the population size is only 1000 individuals. However, it appears that the model becomes less stable when the population size is below about 2000. With smaller populations, there is more risk that the mutation rate could fall to zero by random chance.

## 6.8    Experiment 8: Varying Number of Bits for Encoding Mutation Rate

In the final experiment, the number of bits in the genome used to encode the mutation rate was varied from 5 to 15. The baseline value of `MUTATION_CODE_LENGTH` was 10. The `RUN_LENGTH` was constant at 1,000,000 and the remaining parameters were set to their baseline values. Figure 9 shows the behaviour of the model, averaged over ten separate runs.

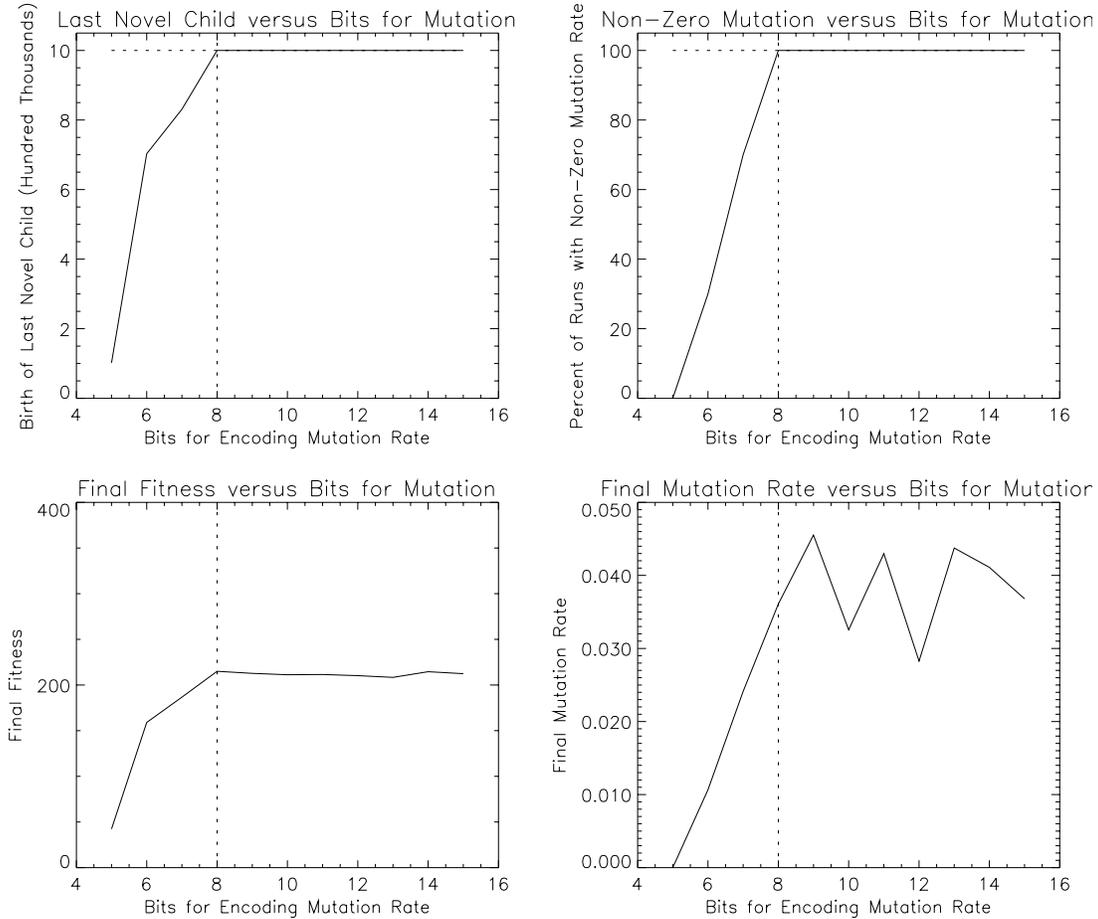

Figure 9: Experiment 8: Varying number of bits for encoding mutation rate. In this experiment, the number of bits in the genome used to encode the mutation rate is varied from 5 to 15. The model displays unbounded evolutionary versatility when the number of bits was more than about 8. When `MUTATION_CODE_LENGTH` is less than 8, it seems that quantization effects make the model unstable. When the encoding is too short, the ideal mutation rate may lie between zero and the smallest value that can be encoded, so the genetic algorithm is forced to set the mutation rate to zero, even though this is less than ideal.

The results suggest that the model will display unbounded evolutionary versatility when the MUTATION_CODE_LENGTH is greater than about 8. If the code length is less than 8 bits, the model becomes susceptible to quantization errors. For example, 6 bits can only encode $2^6 = 64$ values. If the ideal mutation rate is between 0 and $1/(2^6 - 1) = 0.015873$, then the genome may be forced to set the mutation rate to zero, although a non-zero value (but less than 0.015873) would be better.

## 7. Implications of the Model

I do not claim that the model shows that there is a large-scale trend towards increasing evolutionary versatility in organismal evolution; I claim that the model *supports* the idea that, under certain conditions, it is *possible* for evolutionary versatility to increase without bound. In this model, there is active selection for increased evolutionary versatility; there is a selective force that drives the increase; it is not merely a statistical artifact, due to a bounded diffusion process. The model shows how a purely *local* selection process can yield a *global* trend.

The model also shows that the environment must be highly dynamic (the target for the fitness function must change significantly and repeatedly) to sustain increasing evolutionary versatility. If the environment is not sufficiently dynamic, the disruptive effects of mutation will outweigh the beneficial effects, and selection will drive the mutation rates to zero. When the mutation rate is zero throughout the population, the genome length can no longer increase, so the evolutionary versatility is bounded by the length of the longest genome in the population.

I believe that, in fact, there *is* a large-scale trend towards increasing evolutionary versatility in organismal evolution. Although the model does not (and cannot) prove this belief, the model suggests a way to test the belief, because the model predicts that, where there is increasing evolutionary versatility, there should be an accelerating pace of evolution (see the fourth plots in Figures 2 and 3). Therefore, I predict that we will find evidence for an accel-erating pace in the evolution of biological organisms.

It is difficult to objectively verify the claim that the pace of evolution is accelerating. The natural measure of the pace of evolution is the historical frequency of innovations, but the analysis is complicated by several factors. One confounding factor is that the record of the recent past is superior to the record of the distant past, which may give the illusion that there are more innovations in the recent past than the distant past. Another confounding factor is population growth. We may expect more innovations in recent history simply because there are more innovators. A third factor is difficulty of counting innovations. There is a need for an objective threshold on the importance of the innovations, so that the vast number of trivial innovations can be ignored.

I suggest some tests that avoid these objections. I predict that the fossil record will show a decreasing recovery time from major catastrophes, such as mass extinction events, ice ages, meteorite impacts, and volcanic eruptions. Also, I predict a decrease in the average lifetimes of species, as they are out-competed by more recent species at an accelerating rate [15]. These two tests do not involve counting the frequency of innovations, which makes them relatively objective.

## 8.  Limitations and Future Work

There are several limitations to this work. One limitation is that we cannot run the model to infinity, so we cannot prove empirically that evolutionary versatility will grow to infinity. I conjecture that, with the baseline settings of the parameters (Table 1), the expected (i.e., mean, average) evolutionary versatility of the model will rise to infinity as RUN_LENGTH rises to infinity. This conjecture can be *supported* by empirical evidence (Figure 3), but it can only be *proven* by theoretical argument. I have not yet developed this theoretical argument.

Another limitation of the model is its abstractness. A more sophisticated model would include (1) a non-trivial genotype-phenotype mapping, (2) an internal, implicit fitness function, instead of the current, external, explicit fitness function, (3) a genotype-phenotype map-ping and fitness function that allow varying degrees of dependence and independence among the dimensions (i.e., traits, characteristics) along which variation can occur in evolution, (4) the possibility of covariation, (5) coevolution, (6) multiple species, (7) predator-prey relationships, and so on. However, the point of this exercise was to make the model as abstract as possible, in order to identify the minimum elements that are needed to display unbounded evolutionary versatility. The abstractness of the model was intended to make it more clear and susceptible to analysis.

There might seem to be some conflict between this model and the "no free lunch" theorems [43]. Informally, the "no free lunch" theorems show that there is no universal optimization algorithm that is optimal for all fitness landscapes. For example, one "no free lunch" theorem (Theorem 1 in [43]) shows that, for any two optimization algorithms $a_1$ and $a_2$, the average fitness obtained by $a_1$ equals the average fitness obtained by $a_2$, when the average is calculated over all possible fitness landscapes, sampled with uniform probability. If my model can reach infinite fitness levels, for some fitness landscapes, does this violate a "no free lunch" theorem, since then the average fitness must also be infinite? There is no problem here, because the "no free lunch" theorems are concerned with the fitness after a finite number of iterations, not with the fitness after an infinite number of iterations (in my case, an infinite number of children).

The model that is presented here is not intended to be a new, superior form of optimization algorithm. The intent of the model is to show that it is possible, under certain conditions, for evolutionary versatility to increase without bound. Furthermore, the model is intended to show that local selection (in this case, local to a certain period of time) can drive a global trend (global across all periods of time) towards increasing evolutionary versatility. The model is not universal; it will only display unbounded increase in evolutionary versatility for certain parameter settings and for certain fitness landscapes. The fitness landscape is defined by the parameters TARGET_CHANGE_RATE and ERA_LENGTH and by the general design of the model (Figure 1). Experiments 1 and 2 show that the model appears to display unbounded evolutionary versatility for the baseline fitness landscape (the fitness landscape that is defined by the parameter settings in Table 1), but experiments 4 and 5 show that there are neighbouring fitness landscapes for which evolutionary versatility is bounded. The experiments here have only explored a few of the infinitely many possible fitness landscapes. Of the fitness landscapes that were explored here, only a few appeared to display unbounded evolutionary versatility.

## 9. Conclusions

This paper introduces a simple model of unbounded evolutionary versatility. The model is primarily intended to address the claim that natural selection cannot produce a large-scale trend, because it is a purely local process. The model shows that local selection can produce a global trend towards increasing evolutionary versatility. The model suggests that this trend can continue without bound, if there is sufficient ongoing change in the environment.

For evolutionary versatility to increase without bound, it must be possible for the lengths of genomes to increase. If there is a bound on the length of the genomes, then there must be a bound on the evolutionary versatility. A model of unbounded evolutionary versatility must therefore allow mutations that occasionally change the length of a genome. It seems possible that, once genomes reach a certain length, the benefit that might be obtained from greater length would be countered by the damage that mutation can do to the useful genes that have been found so far. At this point, evolutionary versatility might stop increasing.

To address this issue, the model allows the mutation rate to adapt. The experiments show that, indeed, if there is little change in the environment, then the damage of mutation is greater than the benefit of mutation, so the mutation rate goes to zero and evolutionary versatility stops increasing. However, if there is sufficient change in the environment, it appears that the mutation rate reaches a stable non-zero value and evolutionary versatility continues to increase indefinitely.

Perhaps the most interesting observation is that the fitness increase during an era grows over time (see the fourth plots in Figures 2 and 3). That is, increasing evolutionary versatility leads to an accelerating pace of evolution. One of the most interesting questions about this model is whether it plausible as a highly abstract model of the evolution of life on earth. One test of its plausibility is to look for signs that the pace of organismal evolution is accelerating. For example, does the fossil record show a decreasing recovery time from major catastrophes? Is there a decrease in the average lifetimes of species?

If there is evidence that the pace of evolution is accelerating, evolutionary versatility may be better able to account for this than the other seven live hypotheses [21]. It is not clear how any of the other hypotheses could be used to explain the acceleration, although it seems to be a natural consequence of increasing evolutionary versatility.

## Acknowledgments

Thanks to the reviewers for their very helpful comments on an earlier version of this paper. Thanks to Dan McShea for many constructive criticisms and general encouragement.

## References

1. Aboitiz, F. (1991). Lineage selection and the capacity to evolve. *Medical Hypotheses,* 36, 155-156.

2. Altenberg, L. (1994). The evolution of evolvability in genetic programming. In: *Advances in Genetic Programming,* K. E. Kinnear Jr., (ed.). MIT Press.

3. Anderson, R.W. (1995). Learning and evolution: A quantitative genetics approach. *Journal of Theoretical Biology,* 175, 89-101.

4. Ayala F.J. (1974). The concept of biological progress. In *Studies in the Philosophy of Biology*, ed. F.J. Ayala, T. Dobzhansky,19:339-55. New York: Macmillan.

5. Ayala F.J. (1988). Can "progress" be defined as a biological concept? In *Evolutionary Progress*, ed. M Nitecki, pp. 75-96. Chicago: University of Chicago Press.

6.  Bäch, T. (1992). Self-adaptation in genetic algorithms. In F.J. Varela and P. Bourgine (eds.), *Towards a Practice of Autonomous Systems.* MIT Press, pp. 263-271.

7.  Baldwin, J.M. (1896). A new factor in evolution. *American Naturalist,* 30, 441-451.

8.  Bedau, M.A., and Seymour, R. (1995). Adaptation of mutation rates in a simple model of evolution. *Complexity International*, 2.

9.  Blickle, T., and Thiele, L. (1995). *A Comparison of Selection Schemes used in Genetic Algorithms.* Technical Report No. 11. Gloriastrasse 35, CH-8092 Zurich: Swiss Federal Institute of Technology (ETH) Zurich, Computer Engineering and Communications Networks Lab (TIK).

10. Blickle, T., and Thiele, L. (1995). A mathematical analysis of tournament selection, In *Proceedings of the Sixth International Conference on Genetic Algorithms, ICGA-95,* L.J. Eshelman, Ed., Morgan Kaufmann, San Mateo, CA, pp. 9-16.

11. Davis, L. (1989). Adapting operator probabilities in genetic search. *Proceedings of the Third International Conference on Genetic Algorithms, ICGA-89,* Morgan Kaufmann, San Mateo, CA, pp. 61-69.

12. Dawkins, R. (1989). The evolution of evolvability. In: *Artificial Life,* C. Langton, (ed.). Addison-Wesley.

13. Dawkins, R. (1996). *Climbing Mount Improbable.* New York: W.W. Norton and Co.

14. Fogel, D.B., Fogel, L.J., and Atmar, J.W. (1991). Meta-evolutionary programming. In R.R. Chen (Ed.), *Proceedings of the Twenty-fifth Asilomar Conference on Signals, Systems, and Computers,* pp. 540-545, California: Maple Press.

15. Gilinsky, N.L., and Good, I.J. (1991). Probabilities of origination, persistence, and extinction of families of marine invertebrate life. *Paleobiology*, 17, 145-166.

16. Gould S.J. (1988). Trends as changes in variance: A new slant on progress and directionality in evolution. *Journal of Paleontology*, 62, 319-29.

17. Gould S.J. (1997). *Full House: The Spread of Excellence from Plato to Darwin.* New

York: Harmony.

18. Hinton, G.E., and Nowlan, S.J. (1987). How learning can guide evolution. *Complex Systems,* 1, 495-502.

19. Lewin, R. (1985). Red Queen runs into trouble? *Science*, 227, 399-400.

20. McShea, D.W. (1996). Metazoan complexity and evolution: Is there a trend? *Evolution*, 50, 477-492.

21. McShea, D.W. (1998). Possible largest-scale trends in organismal evolution: Eight "Live Hypotheses". *Annual Review of Ecology and Systematics,* 29, 293-318.

22. Nitecki, M.H. (1988). *Evolutionary Progress.* Edited collection. Chicago: University of Chicago Press.

23. Raup, D.M. (1975). Taxonomic survivorship curves and Van Valen's Law. *Paleobiology*, 1, 82-86.

24. Riedl, R. (1977). A systems-analytical approach to macro-evolutionary phenomena. *Quarterly Review of Biology*, 52, 351-370.

25. Riedl, R. (1978). *Order in Living Organisms: A Systems Analysis of Evolution.* Translated by R.P.S. Jefferies. Translation of *Die Ordnung Des Lebendigen.* New York: Wiley.

26. Ruse, M. (1996). *Monad to Man: The Concept of Progress in Evolutionary Biology.* Massachusetts: Harvard University Press.

27. Simon, H.A. (1962). The architecture of complexity. *Proceedings of the American Philosophical Society,* 106, 467-482.

28. Syswerda, G. (1989). Uniform crossover in genetic algorithms. *Proceedings of the Third International Conference on Genetic Algorithms (ICGA-89),* pp. 2-9. California: Morgan Kaufmann.

29. Syswerda, G. (1991). A study of reproduction in generational and steady-state genetic algorithms. *Foundations of Genetic Algorithms.* G. Rawlins, editor. Morgan Kaufmann. pp. 94-101.

30. Turney, P.D. (1989). The architecture of complexity: A new blueprint. *Synthese,* 79, 515-542.

31. Turney, P.D. (1996). How to shift bias: Lessons from the Baldwin effect. *Evolutionary Computation,* 4, 271-295.

32. Turney, P.D. (1999). Increasing evolvability considered as a large-scale trend in evolution. In A. Wu, ed., *Proceedings of 1999 Genetic and Evolutionary Computation Conference Workshop Program (GECCO-99 Workshop on Evolvability),* pp. 43-46.

33. Van Valen, L. (1973). A new evolutionary law. *Evolutionary Theory,* 1, 1-30.

34. Vermeij, G.J. (1970). Adaptive versatility and skeleton construction. *American Naturalist,* 104, 253-260.

35. Vermeij, G.J. (1971). Gastropod evolution and morphological diversity in relation to shell geometry. *Journal of Zoology,* 163, 15-23.

36. Vermeij, G.J. (1973). Biological versatility and earth history. *Proceedings of the National Academy of Sciences of the United States of America,* 70, 1936-1938.

37. Vermeij, G.J. (1974). Adaptation, versatility, and evolution. *Systematic Zoology,* 22, 466-477.

38. Wagner, G.P. and Altenberg, L. (1996). Complex adaptations and the evolution of evolvability. *Evolution,* 50, 967-976.

39. Whitley, D., and Kauth, J. (1988). GENITOR: A different genetic algorithm. *Proceedings of the Rocky Mountain Conference on Artificial Intelligence,* Denver, CO. pp. 118-130.

40. Whitley, D. (1989). The GENITOR algorithm and selective pressure. *Proceedings of the Third International Conference on Genetic Algorithms (ICGA-89),* pp. 116-121. California: Morgan Kaufmann.

41. Whitley, D., Dominic, S., Das, R., and Anderson, C.W. (1993). Genetic reinforcement learning for neurocontrol problems. *Machine Learning,* 13, 259-284.

42. Williams, G.C. (1966). *Adaptation and Natural Selection.* New Jersey: Princeton University Press.

43. Wolpert, D.H., and Macready, W.G. (1997). No free lunch theorems for optimization. *IEEE Transactions on Evolutionary Computation*, 1, 67-82.